\documentclass[letterpaper]{article} % DO NOT CHANGE THIS
\usepackage{aaai24}  % DO NOT CHANGE THIS
\usepackage{times}  % DO NOT CHANGE THIS
\usepackage{helvet}  % DO NOT CHANGE THIS
\usepackage{courier}  % DO NOT CHANGE THIS
\usepackage[hyphens]{url}  % DO NOT CHANGE THIS
\usepackage{graphicx} % DO NOT CHANGE THIS
\usepackage{natbib}  % DO NOT CHANGE THIS AND DO NOT ADD ANY OPTIONS TO IT
\usepackage{caption} % DO NOT CHANGE THIS AND DO NOT ADD ANY OPTIONS TO IT
\usepackage{algorithm}
\usepackage{algorithmic}

\usepackage{amsthm,amsmath,amssymb}
\usepackage{mathrsfs}
\usepackage{bm}

\usepackage{newfloat}
\usepackage{listings}
\DeclareCaptionStyle{ruled}{labelfont=normalfont,labelsep=colon,strut=off} % DO NOT CHANGE THIS
\floatstyle{ruled}
\newfloat{listing}{tb}{lst}{}
\floatname{listing}{Listing}
\title{Causal Walk: Debiasing Multi-Hop Fact Verification with Front-Door Adjustment}
\author{
    Congzhi Zhang\equalcontrib,
    Linhai Zhang\equalcontrib,
    Deyu Zhou\thanks{~Corresponding author.}
}
\affiliations{
    School of Computer Science and Engineering, Key Laboratory of Computer Network \\
    and Information Integration, Ministry of Education, Southeast University, China \\
    \{zhangcongzhi, lzhang472, d.zhou\}@seu.edu.cn}
\usepackage{bibentry}
\begin{document}

\maketitle

\begin{abstract}
Multi-hop fact verification aims to detect the veracity of the given claim by integrating and reasoning over multiple pieces of evidence. 
Conventional multi-hop fact verification models are prone to rely on spurious correlations from the annotation artifacts, leading to an obvious performance decline on unbiased datasets. 
Among the various debiasing works, the causal inference-based methods become popular by performing theoretically guaranteed debiasing such as casual intervention or counterfactual reasoning. 
However, existing causal inference-based debiasing methods, which mainly formulate fact verification as a single-hop reasoning task to tackle shallow bias patterns, cannot deal with the complicated bias patterns hidden in multiple hops of evidence. 
To address the challenge, we propose Causal Walk, a novel method for debiasing multi-hop fact verification from a causal perspective with front-door adjustment. 
Specifically, in the structural causal model, the reasoning path between the treatment (the input claim-evidence graph) and the outcome (the veracity label) is introduced as the mediator to block the confounder.  
With the front-door adjustment, the causal effect between the treatment and the outcome is decomposed into the causal effect between the treatment and the mediator, which is estimated by applying the idea of random walk, and the causal effect between the mediator and the outcome, which is estimated with normalized weighted geometric mean approximation. 
To investigate the effectiveness of the proposed method, an adversarial multi-hop fact verification dataset and a symmetric multi-hop fact verification dataset are proposed with the help of the large language model. 
Experimental results show that Causal Walk outperforms some previous debiasing methods on both existing datasets and the newly constructed datasets. Code and data will be released at https://github.com/zcccccz/CausalWalk.
\end{abstract}

\section{Introduction}

Fact verification aims to verify the given claim based on the retrieved evidence, which is a challenging task. 
Previous work formulates fact verification as a natural language inference task, where multiple evidence pieces are concatenated together and a single-hop inference is performed~\cite{hanselowski-etal-2018-ukp, nie2019combining}. 
However, in many cases, the process of verifying a claim requires integrating and reasoning over several pieces of evidence~\cite{ijcai2021p536}. 
Therefore, multi-hop fact verification, which performs a multi-hop reasoning process to verify a claim, has become an attractive research topic recently~\cite{zhou-etal-2019-gear, Zhao2020Transformer-XH:, si2023exploring}.

\begin{figure}[t]
\centering
\includegraphics[width=0.47\textwidth]{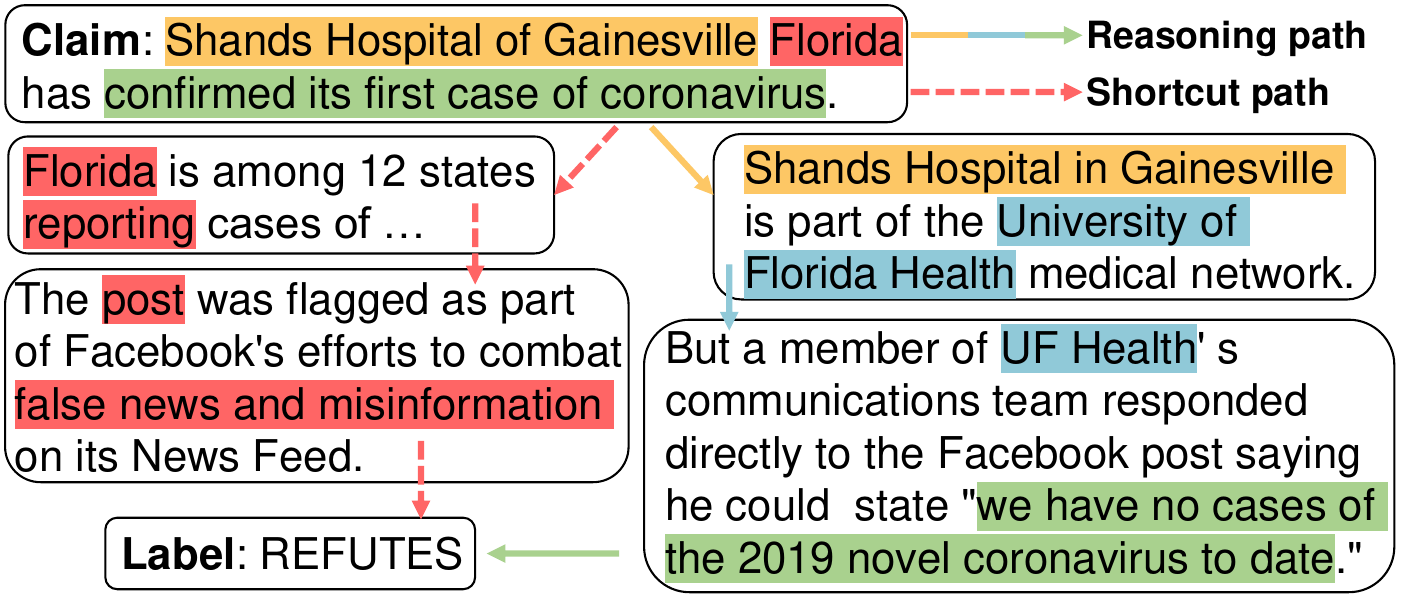}
\caption{Illustration of an example of bias in multi-hop fact verification dataset, which is taken from the PolitiHop dataset. The solid line indicates the reasoning path while the dashed line indicates the shortcut path.} 
\label{fig:example}
\end{figure}

Though notable progress has been made, most multi-hop fact verification methods focus on learning label-specific features for judging the veracity of the given claim, which may expose the models to hidden data bias.
Previous study~\cite{schuster-etal-2019-towards} has shown that there are annotation biases in a commonly used fact verification dataset, FEVER~\cite{thorne-etal-2018-fever}, where most of the claims require only a single piece of evidence to verify. 
As shown in Figure~\ref{fig:example}, we observe that there are also biases in the multi-hop fact verification datasets such as PolitiHop~\cite{ijcai2021p536}. 
The sentence ``The post was flagged as part ...'' appears 85 times in the training set of PolitiHop with a high correlation with the REFUTES label. 
Therefore, it is easy for the black-box neural network methods to learn such shortcut paths instead of multi-hop reasoning to cheat and obtain the right answer.
We found obvious performance declines for some popular multi-hop fact verification methods when such biases are removed in a synthetic dataset.
The same phenomenon, known as disconnected reasoning, has also been found in multi-hop question answering datasets~\cite{trivedi-etal-2020-multihop}.

Previous methods for debiasing cannot deal with biases in multi-hop datasets as they mainly formulate fact verification as a single-hop reasoning task.  
For data augmentation-based methods~\cite{wei-zou-2019-eda, 10.1145/3459637.3482078}, it is difficult to generate unbiased multi-hop fact verification instances.
For reweight-based methods~\cite{schuster-etal-2019-towards, karimi-mahabadi-etal-2020-end}, it is hard to detect the biased samples as the bias patterns are complicated.
Recently, causal inference has become a popular paradigm for debiasing methods because of its theoretical guarantee and generalizability, which aims to calculate the causal effect between the treatments (input examples) and outputs (labels).
For causal inference-based methods, the common practices are causal intervention~\cite{Tian_Cao_Zhang_Xing_2022} and counterfactual reasoning~\cite{xu-etal-2023-counterfactual}. 
However, these methods mainly focus on shallow bias patterns such as the correlation between specific types of words (e.g. negation words) and specific labels (e.g. REFUTES), and cannot deal with complicated bias patterns hidden in multiple hops of evidence. 

\begin{figure}[t]
\centering
\includegraphics[width=0.46\textwidth]{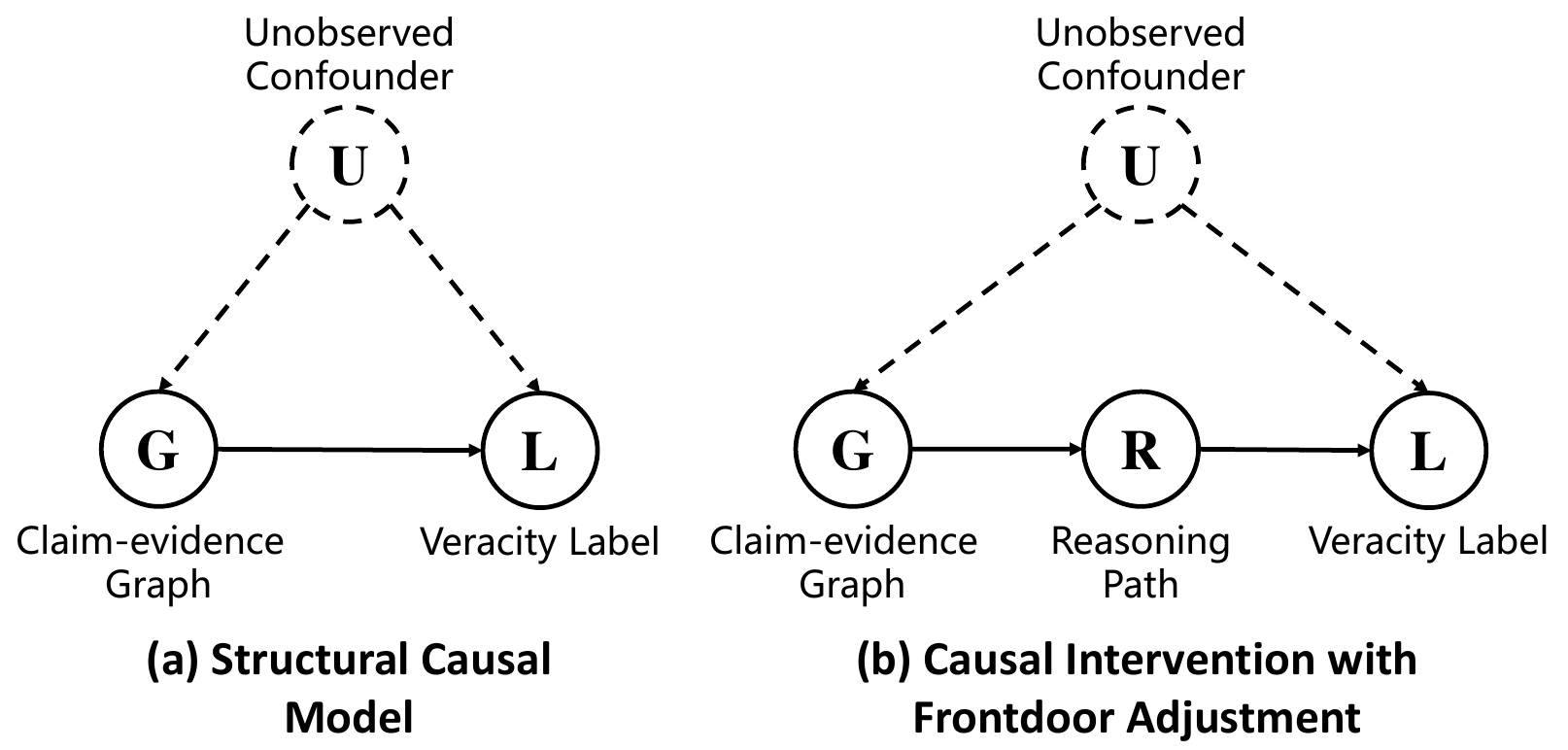}
\caption{Structural Causal Model for multi-hop fact verification.} 
\label{fig:SCM}
\end{figure}

To address the above challenge, we propose to debias the multi-hop fact verification by causal intervention based on front-door adjustment. 
As shown in Figure~\ref{fig:SCM}(a), we reflect the causal relationships in the multi-hop fact verification as a Structural Causal Model (SCM), where $G$ is the graph consisting of claim and evidence. $U$ is the unobservable confounder that introduces various biases. $R$ is the reasoning path. $L$ is the corresponding veracity label.
The debiasing process is achieved by measuring the causal effect between treatment $G$ and outcome $L$. 
As $U$ absorbs various multi-hop biases, making it hard to model or detect, it becomes infeasible to employ back-door adjustment to calculate the causal effect between $G$ and $L$.
As shown in Figure~\ref{fig:SCM}(b), we introduce the reasoning path between $G$ and $L$ as a mediator variable $R$, which is unaffected by $U$ and fully mediates the causal effect between $G$ and $L$. 
With $R$ introduced, the front-door adjustment can be employed by measuring the causal effect between $G$ and $L$ as an adding up of the causal effect between $G$ and $R$ and the causal effect between $R$ and $L$.

In this paper, Causal Walk, a novel debiasing method for multi-hop fact verification based on front-door adjustment, is proposed. 
Specifically, to measure the causal effect between $G$ and $R$, the idea of random walk is applied with the graph neural network to calculate the probability of the reasoning path.
To measure the causal effect between $R$ and $L$, normalized weighted geometric mean (NWGM) approximation is utilized with the recurrent neural network to estimate the unbiased outcome based on the reasoning path.
Furthermore, we construct an adversarial multi-hop fact verification dataset and a symmetric multi-hop fact verification dataset by extending the PolitiHop dataset with the help of large language models. 
We conduct experiments on both single-hop and multi-hop fact verification datasets under both original and adversarial scenarios. 
Experimental results show that the proposed method outperforms previous debiasing methods on both single-hop and multi-hop datasets.
The contributions of this work are three-fold.
\begin{itemize}
\item As far as we know, we are the first one to debias multi-hop fact verification task using front-door adjustment.
\item We propose Causal Walk, a novel method to perform causal intervention with front-door adjustment by introducing the reasoning path between input and output as the mediator.
\item Experimental results show the effectiveness of the proposed method on both previous datasets and the proposed datasets.
\end{itemize}

\section{Related Work}

\subsection{Multi-hop Fact Verification}

Early methods of fact verification mainly formulate fact verification as a natural language inference task~\cite{hanselowski-etal-2018-ukp, nie2019combining}. 
\citet{zhou-etal-2019-gear} first propose a graph-based evidence reasoning framework to enable the communication of multiple pieces of evidence in a fully connected graph.
\citet{zhong-etal-2020-reasoning} introduce a semantic-level graph for fact verification.
\citet{liu-etal-2020-fine} utilize the Kernel Graph Attention Network for fine-grained fact verification.
\citet{Zhao2020Transformer-XH:} update Transformer~\cite{NIPS2017_3f5ee243} with extra Hop attention for multi-hop reasoning tasks including fact verification.
\citet{10.1145/3485447.3512135} propose an evidence fusion network to capture global contextual information from various levels of evidence information.
\citet{si2023exploring} cast explainable multi-hop fact verification as subgraph extraction with salience-aware graph learning.
\citet{fajcik-etal-2023-claim} present a 2-stage system composed of the retriever and the verifier, while we only focus on the debiasing of the verifier. Recently, some works have used LLM for multi-hop fact verification~\cite{zeng-gao-2023-prompt,pan-etal-2023-fact}.
Most of the existing multi-hop fact verification methods focus on modeling the reasoning process while ignoring the hidden bias in the datasets.

\subsection{Debiasing with Causal Inference}
Recently, causal inference has been preferred for the debiasing method, which provides a more principled way of defining a causal model and debiasing the model by measuring the causal effect. 
\citet{xu-etal-2023-counterfactual} propose to mitigate the spurious correlation between the claims and the labels by subtracting the output of a claim-only model from the output of a claim-evidence fusion model.
\citet{Tian_Cao_Zhang_Xing_2022} combine the causal intervention and counterfactual reasoning to debias fact verification, where the do-calculus for causal intervention is estimated by NWGM approximation.
Besides fact verification, causal inference is also widely applied in debiasing other Natural Language Processing tasks and Computer Vision tasks. 
\citet{wang-etal-2022-causal} introduced instrumental variable estimation to debias implicit sentiment analysis. 
\citet{guo-etal-2023-counterfactual} employed counterfactual reasoning for reducing disconnected reasoning in multi-hop QA.
\citet{niu2021counterfactual} proposed to use counterfactual reasoning to debias the visual question answering task by subtracting the prediction of the language-only model from the prediction of the vision-language model.
Some works apply backdoor adjustment or front-door adjustment to the image caption task~\cite{Liu_2022_CVPR1,yang2021causal,yang2021deconfounded}.
\citet{zhu-etal-2023-causal} proposed neuron-wise and token-wise backdoor adjustments to mitigate name bias in machine reading comprehension.
\citet{chen-etal-2023-causal} proposed to debias multi-modal fake news detection with both causal intervention and counterfactual reasoning. 
To our best knowledge, we are the first one to utilize front-door adjustment for multi-hop fact verification debiasing.

\section{Preliminaries}
\subsection{Structural Causal Model and Causal Effect}
The structural Causal Model reflects the causal relationships between certain variables we are interested in. 
As shown in Figure~\ref{fig:SCM}(a), SCM is often represented as a directed acyclic graph $ SCM = \{V, E\}$, where $V$ denotes the set of variables and $E$ denotes the direct causal effect.
$G$ is a direct cause of $L$ when variable $L$ is the child of $G$.
As for fact verification, we regard the graph that combines the claim and corresponding evidence as the treatment variable $G$, and the veracity label as the outcome variable $L$. 
The claim and evidence together determine the veracity label. 
Therefore we have an edge $G \rightarrow L$ to show the direct causal effect of $G$ on $L$.
Except for the input and output variables, there is another unobservable variable that affects both the input and output as a background, which we denote as the unobservable confounder variable $U$.
For fact verification, $U$ can be annotation artifacts that inevitably introduce biases between the input and output. 
For example, most annotators will follow certain patterns including negation words when generating REFUTES instances. 
So we have another path $G \leftarrow U \rightarrow L$ from $G$ to $L$, which is also known as a backdoor path. 

The conventional methods often adopt the total effect $P(L|G)$ to measure how input $G$ affects output $L$.
However, from the perspective of causal inference, such total effect does not reflect the real effect of $G$ on $L$, it involves all paths from $G$ to $L$ including the backdoor path.
Therefore, methods based on $P(L|G)$ have poor generalizability on out-of-domain datasets and are vulnerable to adversarial attacks for learning biases hidden in the backdoor path unconsciously.
Contrary to conventional methods, causal inference calculates the direct causal effect between $G$ and $L$ with do-calculus $P(L|do(G))$.
The do-calculus $P(L|do(G))$ measures the effect on $L$ when intervening the treatment $G$ but keeping other variables unchanged, while $P(L|G)$ only means the probability of $L$ condition on $G$.

\subsection{Causal Intervention with Front-door Adjustment}
There are four main ways to calculate the do-calculus: randomized controlled trial, backdoor adjustment, front-door adjustment, and instrumental variable estimation. 
Both the randomized controlled trial method and the instrumental variable estimation method require a thorough control of the relationship between input text and label, which is difficult for multi-hop fact verification. 
The backdoor adjustment is also infeasible as the confounder $U$ is too polymorphic to be observed exhaustively.
Therefore, the front-door adjustment is chosen to perform the causal intervention and calculate $P(L|do(G))$.

To perform the front-door adjustment, a mediator variable is required to fully mediate the causal effect between $G$ and $L$, i.e. all direct causal paths from $G$ to $L$ go through the mediator.
For multi-hop fact verification, we choose the reasoning path in the claim-evidences graph as the mediator variable $R$. 
On the one hand, $R$ is only affected by the graph $G$ itself, on the other hand, the only way the graph $G$ can affect the label $L$ is through the reasoning path $R$.
Therefore, we can specify the path $G \rightarrow R \rightarrow L$.
With the mediator, based on the front-door adjustment, we have
\begin{equation}
P(L|do(G)) = \sum_{r}P(L|do(r))P(r|do(G))
\label{front-door}
\end{equation}
where $r \in R$ is the reasoning path between $G$ and $L$. The causal effect between $G$ and $L$ is decomposed into the causal effect between $G$ and $R$ and the causal effect between $R$ and $L$.

Since $L$ is a collider of $G$ and $R$, $L$ blocks the backdoor paths of $G$ and $R$. Therefore, based on the backdoor adjustment, we have
\begin{equation}
P(r|do(G)) = \sum_{l}P(r|G,l)P(l)=P(r|G)
\label{P(r|do(G))}
\end{equation}
where $l \in L$ is the veracity label.

Because $G$ blocks $R\leftarrow G\rightarrow U\rightarrow L$, $G$ satisfies the backdoor criterion, and we have
\begin{equation}
P(L|do(r)) = \sum_{g}P(L|r,g)P(g)
\label{P(L|do(r))}
\end{equation}
where $g \in G$ is the claim-evidence graph.

\begin{figure*}[htbp]
 \centering
 \includegraphics[width=1.0\textwidth]{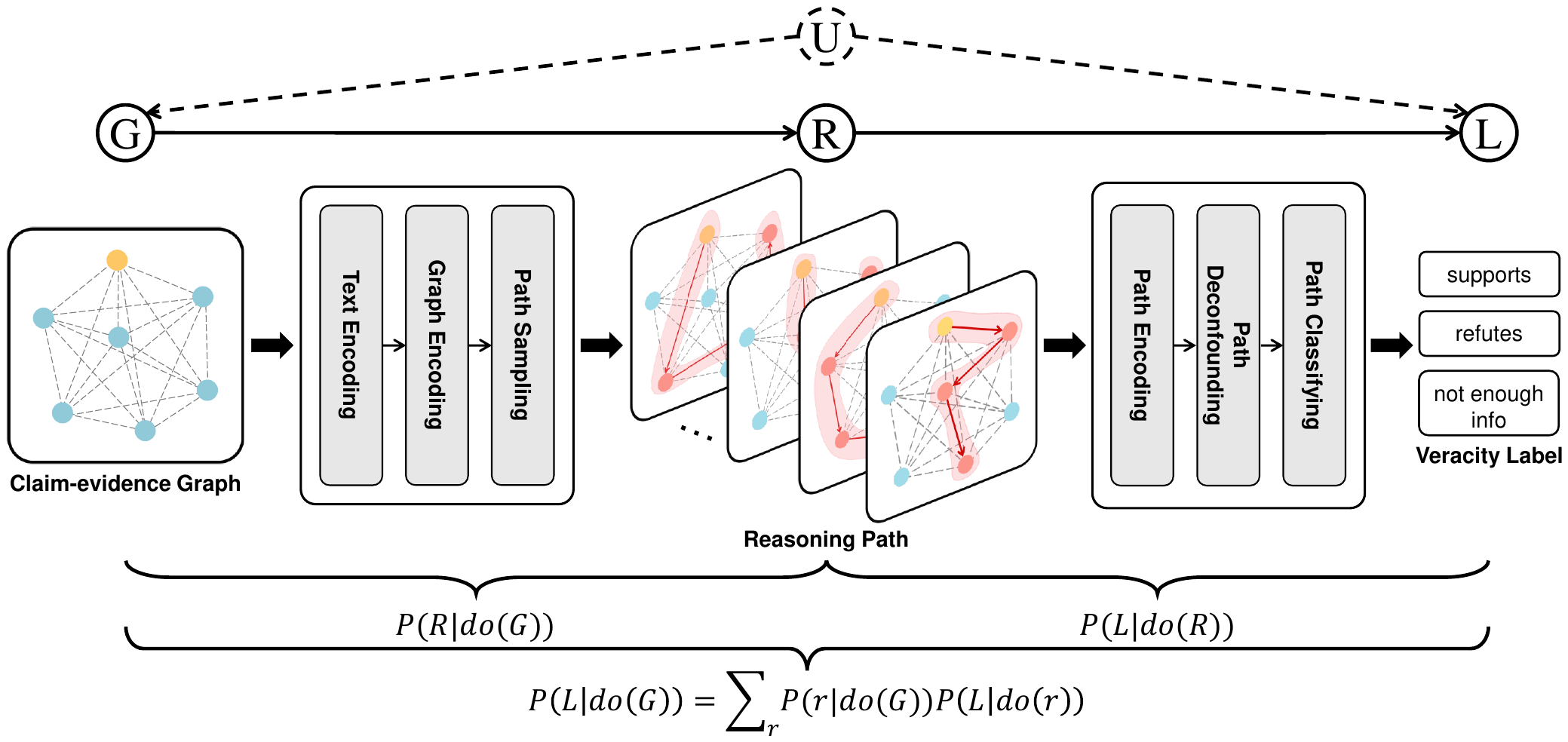}
 \caption{The causal view of the proposed method. }
 \label{fig:model}
\end{figure*} 

\section{Methodology}

In this section, multihop fact verification is formulated as a graph-based classification task. We first estimate $P(r|do(G))$ by applying the idea of random walk with the graph neural network. We then combine Normalized Weighted Geometric Mean approximation with the recurrent neural network to estimate $P(L|do(r))$. Finally, we use the beam search to estimate $P(L|do(G))$ and train the model.

\subsection{Task Definition}
Given the claim $c$ and the corresponding evidence set $\{e_1,...,e_n\}$, the model needs to detect the veracity of the claim. Following the previous work~\cite{zhou-etal-2019-gear}, we combine the claim and evidence set into a graph $G$. Each node in graph $G$ represents a sentence. We define multi-hop fact verification as a graph classification problem, with labels including SUPPORTS, REFUTES, and NOT ENOUGH INFO. Specifically, we define the graph $G=\{v_0,... v_n\}$. $G$ can be divided into two subgraphs $G_{claim}=\{v_0\}$ and  $G_{evidence}=\{v_1,... v_n\}$. Since there is no more fine-grained prior information, we set graph $G$ to be fully connected.

% Specifically, we define the graph $G=\{\bm{A},\bm{X}\}$, which contains a node set $V$ and edge set $E$, where $\bm{X}\in \mathbb{R}^{|V|\times F}$ represents the nodes feature matrix, where $\bm{x}_i=\bm{X}[i,:]$ is an $F$-dimensional feature vector of node $v_i\in V$. $\bm{A}\in \mathbb{R}^{|V|\times|V|}$ is an adjacency matrix representing the connection information of nodes in the graph. If there is an edge between the node $v_i$ and $v_j$, i.e. $(v_i,v_j)\in E$, then $\bm{A}[i,j]=1$, otherwise $\bm{A}[i,j]=0$.

\subsection{Estimation of $\bm{P(r|do(G))}$ }
In this subsection, we will introduce how to use the walk-based method to estimate the $P(r|do(G))$ in the Equation (\ref{front-door}). 
According to the Equation (\ref{P(r|do(G))}), the estimation of $P(r|do(G))$ is equivalent to the estimation of $P(r|G)$. 
To estimate the probability $P(r|G)$, we first obtain the node representation using the \textbf{Text Encoding} and \textbf{Graph Encoding}. 
We then compute the transition probability matrix using the node representation. 
Finally, we sample a path $r$ with probability $P_{walk}(r)$ through the \textbf{Path Sampling}. 

\subsubsection{Text Encoding}
Following the previous work~\cite{zhou-etal-2019-gear,liu-etal-2020-fine}, we employ BERT~\cite{devlin-etal-2019-bert} to obtain the semantic representation of the text.

The claim is directly fed into BERT to obtain the claim representation $\bm{x}^{c}$ while the evidence is concatenated with the claim as an evidence-claim pair $(e_i,c)$ into BERT to obtain the evidence representation $\bm{x}_{i}^{e}$. 
\begin{equation}
    \begin{aligned}
        \bm{x}_{0}^{c}&={\rm BERT}(c) \\
        \bm{x}_{i}^{e}&={\rm BERT}(e_i,c) 
    \end{aligned}
    \label{bert}
\end{equation}
For simplification, the superscripts of $\bm{x}_{0}^{c}$ and $\bm{x}_{i}^{e}$ are omited in the following paper.

\subsubsection{Graph Encoding}

To capture the structural information of the graph, we use graph convolution to update the node representation.
\begin{equation}
    \begin{aligned}
        \bm{X}^{layer}={\rm GConv}(\bm{X},\bm{A})
    \end{aligned}
\end{equation}
$\bm{X}=\{\bm{x}_{0},\bm{x}_{1},...,\bm{x}_{n}\}\in \mathbb{R}^{|V|\times F}$ is the initialized nodes representation from BERT. $\bm{A}\in \mathbb{R}^{|V|\times|V|}$ is the adjacency matrix. $layer$ represents the number of stacked layers of the graph convolution. $\bm{X}^{layer}\in \mathbb{R}^{|V|\times d}$ represents the updated nodes representation. The following operations in this paper are based on the updated node representation, i.e. $\bm{x}_i=\bm{X}^{layer}[i,:]$.

\subsubsection{Path Sampling}
To sample a path and estimate the probability, we apply the idea of a random walk~\cite{li-etal-2016-discriminative,christopoulou-etal-2018-walk}. The path on the graph represents the inference process for the claim. 

The core parameter of the random walk is the transition probability. We calculate the weights of the adjacency matrix first and then use the softmax function to calculate the transition probability. In a random walk starting from $v_0$, the weight of edge $(v_i,v_j)$ is calculated as follows:
\begin{equation}
    \begin{aligned}
        a_{ij}={\rm MLP}(\bm{x}_i,\bm{x}_j,\bm{x}_0)
    \end{aligned}
\end{equation}

We assume that the random walk follows a first-order Markov chain. The probability of jumping from node $v_i$ to $v_j$ is:
\begin{equation}
    \begin{aligned}
        P(i\rightarrow j)={\rm softmax}(a_{ij})=\frac{exp(a_{ij})}{\sum_{k\in N(i)}exp(a_{ik})}
    \end{aligned}
    \label{transition}
\end{equation}
where $N(i)$ is the neighbor set of node $v_i$.

Then we random sample an inference path $r=\{v_0,...,v_m\}$. Its length is $m+1$. The probability of this path can be calculated by:
\begin{equation}
    \begin{aligned}
        P_{walk}(r)=\prod \limits_{k=0}^{m-1}P(k\rightarrow k+1)
    \end{aligned}
    \label{path_probability}
\end{equation}

According to the assumption of a first-order Markov chain, we consider the conditional probability $P(r|G)$ and the path sampling probability $P_{walk}(r)$ to be equivalent. And then according to the Equation (\ref{P(r|do(G))}), we have
\begin{equation}
    \begin{aligned}
        P(r|do(G))=P(r|G)=P_{walk}(r)
    \end{aligned}
    \label{walk}
\end{equation}

\subsection{Estimation of $\bm{P(L|do(r))}$}
In this subsection, we describe how to use Normalized Weighted Geometric Mean approximation to estimate $P(L|do(r))$. According to the Equation (\ref{P(L|do(r))}), we need to estimate $\sum_{g}P(g)P(L|r,g)$. We first obtain the path representation $\bm{x}_r$ and graph representation $\bm{x}_g$ by utilizing the \textbf{Path Encoding}. In \textbf{Path Deconfounding}, to reduce the computational cost, we use NWGM to absorb the sampling of $g$ into the model, as shown in Equation (\ref{nwgm1}). Since it is not possible to exhaust the values of $g$, we use a fixed dictionary $\bm{D}_g$ to store the compressed representation space of graph $g$. Finally, in \textbf{Path Classifying}, we use a classifier to classify the path representation after causal intervention.

\subsubsection{Path Encoding}
We encode the nodes on the path $r$ with a recurrent neural network to get the path representation $\bm{x}_r$,
\begin{equation}
    \begin{aligned}
        \bm{x}_r={\rm LSTM}([\bm{x}_0,...,\bm{x}_m],\bm{h}_0,\bm{c}_0)
    \end{aligned}
\end{equation}
where $\bm{h}_0$ and $\bm{c}_0$ are the initial state of the LSTM network. We set $\bm{h}_0=\bm{c}_0=\bm{x}_g$. The $\bm{x}_g$ is the graph representation:
\begin{equation}
    \begin{aligned}
        \bm{x}_g={\rm Attention}(\bm{x}_0,[\bm{x}_1,...,\bm{x}_n])
    \end{aligned}
    \label{graph_rep}
\end{equation}
where $n$ is the number of evidence sentences.

Here we use the attention mechanism to learn graph representation:
\begin{equation}
    \begin{aligned}
        w_i&={\rm MLP}(\bm{x}_0,\bm{x}_i) \\
        \alpha_i&={\rm softmax}(w_i)=\frac{exp(w_i)}{\sum_{k=1}^nexp(w_k)} \\
        \bm{x}_g&=\sum_{i=1}^n\alpha_i\bm{x}_i
    \end{aligned}
\end{equation}

\subsubsection{Path Deconfounding}
To estimate $P(L|do(r))$, given the path $r$'s representation $\bm{x}_r$ and graph $g$'s representation $\bm{x}_g$, Equation (\ref{P(L|do(r))}) is implemented as:
\begin{equation}
    \begin{aligned}
        \sum_{g\in G}P(\bm{x}_g)P(l|\bm{x}_r,\bm{x}_g)=\mathbb{E}_g[P(l|\bm{x}_r,\bm{x}_g)]
    \end{aligned}
\end{equation}
where $G$ are the value spaces of $g$. $P(l|\bm{x}_r,\bm{x}_g)$ is the prediction results of classifier $f_{classifier}$:
\begin{equation}
    \begin{aligned}
        P(l|\bm{x}_r,\bm{x}_g)&=f_{classifier}(\bm{x}_r,\bm{x}_g)\\
        &={\rm softmax}(h(\bm{x}_r,\bm{x}_g))
    \end{aligned}
\end{equation}
where $h$ is a feature fusion function.

Note that $\mathbb{E}_g$ cannot be calculated analytically, we use Normalized Weighted Geometric Mean (NWGM)~\cite{Xu_Ba_Kiros_Cho_Courville_Salakhudinov_Zemel_Bengio_2015} to make an approximation of expectation:
\begin{equation}
    \begin{aligned}
        \mathbb{E}_g[P(l|\bm{x}_r,\bm{x}_g)]&=\mathbb{E}_g[{\rm softmax}(h(\bm{x}_r,\bm{x}_g))]\\
        % &\mathop{\approx}\limits^{NWGM} softmax(\mathbb{E}_g[h(x_r,x_g)])
        &\approx {\rm softmax}(\mathbb{E}_g[h(\bm{x}_r,\bm{x}_g)])
    \end{aligned}
    \label{nwgm1}
\end{equation}

Following recent works~\cite{Tian_Cao_Zhang_Xing_2022,chen-etal-2023-causal}, we model $h(\bm{x}_r,\bm{x}_g)=\bm{W}_r\bm{x}_r + \alpha \bm{W}_g\bm{x}_g$, where $\bm{W}_r$ and $\bm{W}_g$ are learnable weight parameters, $\alpha$ is weight parameter for intervention. In this case, $\mathbb{E}_g[h(\bm{x}_r,\bm{x}_g)]=\bm{W}_r\bm{x}_r + \alpha \bm{W}_g\cdot\mathbb{E}_g[\bm{x}_g]$.

Since the values of $g$ are inexhaustible, we propose to approximate it by designing a fixed dictionary $\bm{D}_g=[\bm{z}_1,...,\bm{z}_N]\in \mathbb{R}^{N\times k\times d}$. $N$ is the number of categories. $k$ is the number of samples for each category. $d$ is the feature dimension of graph representation. We couple the dictionary size and the category of labels so that $\bm{D}_g$ can be modeled as the confounder for each category. We use the graph representations from the training dataset to initialize the dictionary $\bm{D}_g$. Specifically, for each category $l_i$, we sample $k$ cluster centers using the K-Means algorithm, i.e. $\bm{z}_i=[\bm{z}_i^1,...,\bm{z}_i^k]\in \mathbb{R}^{k\times d}$, where $\bm{z}_i^k$ is a graph representation and it can be calculated by Equation (\ref{graph_rep}).

To compute $\mathbb{E}_g[\bm{x}_g]$, we use a dot-product attention mechanism:
\begin{equation}
    \begin{aligned}
        \bm{z}_i^{'} &={\rm softmax}(\bm{Q}^T\bm{K})\bm{z}_i \\
        \bm{D}_g^{'}&=[\bm{z}_1^{'},...,\bm{z}_N^{'}] \\
        \mathbb{E}_g[\bm{x}_g]&=\frac{1}{N}P(l|\bm{x}_r)\bm{D}_g^{'}
    \end{aligned}
    \label{dict}
\end{equation}
where $\bm{Q}=\bm{W}_q\bm{x}_r$, $\bm{K}=\bm{W}_k\bm{z}_i^T$ ($\bm{W}_q\in \mathbb{R}^{d\times d}$ and $\bm{W}_k\in \mathbb{R}^{d\times d}$ are learnable mapping matrices). $\bm{z}_i^{'}\in \mathbb{R}^d$ is the graph representation of category $l_i$ computed by the dot-product attention. $\bm{D}_g^{'}\in \mathbb{R}^{N\times d}$ is the confounder dictionary after attention. $\bm{x}_r\in \mathbb{R}^d$ is the path $r$'s representation. 

\subsubsection{Path Classifying}
$P(l|\bm{x}_r)$ in Equation (\ref{dict}) is the prediction results of classifier $f_{classifier}$ given only $r$ input:
\begin{equation}
    \begin{aligned}
        \bm{l}_r=P(l|\bm{x}_r)=f_{classifier}(\bm{x}_r)
    \end{aligned}
    \label{path_class}
\end{equation}
where $\bm{l}_r\in \mathbb{R}^{N}$ is the probability distribution of the classification results.

In summary, with the Normalized Weighted Geometric Mean, we can deduce that
\begin{equation}
    \begin{aligned}
        P(L|do(r))={\rm softmax}(\bm{W}_r\bm{x}_r+\alpha\bm{W}_g\cdot \mathbb{E}[\bm{x}_g])
    \end{aligned}
    \label{nwgm}
\end{equation}

\subsection{Training and Inference}
In the training stage, since we cannot determine which path is the true inference path, we use the beam search to sample several paths with the highest probability. Specifically, we sample a path set $R_{beam}=\{r_1,...,r_w\}$, $w$ is the beam search width. 

According to Equations (\ref{front-door})(\ref{walk})(\ref{nwgm}), we can implement front-door adjustment as:
\begin{equation}
    \begin{aligned}
        \bm{l}_{causal}&=P(L|do(G))\\
        &=\sum_{r}P_{walk}(r)\sum_{g}P(\bm{x}_g)P(L|\bm{x}_r,\bm{x}_g)\\
        &=\mathbb{E}_{r\in R_{beam}}[{\rm softmax}(\bm{W}_r\bm{x}_r+\alpha\bm{W}_g\cdot \mathbb{E}[\bm{x}_g])]
    \end{aligned}
\end{equation}
where $\bm{l}_{causal}\in \mathbb{R}^{N}$ is the probability distribution of the classification results.

Finally, we use the cross-entropy loss for training:
\begin{equation}
    \begin{aligned}
        \mathcal{L}_{causal}=-\bm{l}_{gold}^Tlog(\bm{l}_{causal})
    \end{aligned}
    \label{walk_loss}
\end{equation}
where $\bm{l}_{gold}\in \mathbb{R}^{N}$ is the ground-truth label.

We use the supervision of ground-truth labels to further enhance the learning of reasoning path:
\begin{equation}
    \begin{aligned}
    \begin{split}
        \bm{l}_{pred}&=\sum_{r\in R_{beam}}P_{walk}(r)\bm{l}_r \\
        \mathcal{L}_{walk}&=-\bm{l}_{gold}^Tlog(\bm{l}_{pred})
    \end{split}
    \end{aligned}
    \label{final_classifier}
\end{equation}

The total training loss is:
\begin{equation}
    \begin{aligned}
        \mathcal{L}_{total} = \mathcal{L}_{walk} + \mathcal{L}_{causal}
    \end{aligned}
    \label{walk_loss}
\end{equation}

\section{Experiments}

\subsection{Datasets}
We evaluate the model performance on the FEVER dataset and PolitiHop dataset and their variants. For all datasets, label classification accuracy is adopted as the evaluation metric. 
For training, all models are trained on the original training set of FEVER and PolitiHop.
For testing, the developed set of FEVER and the test set of PolitiHop are adopted, denoted as \textbf{FEVER}~\cite{thorne-etal-2018-fever} and \textbf{PolitiHop}~\cite{ijcai2021p536} respectively. 
We also include the adversarial versions of both datasets, denoted as \textbf{Adversarial FEVER}~\cite{thorne-etal-2018-fever} and \textbf{Adversarial PolitiHop}~\cite{ijcai2021p536} respectively. 
To further investigate the effectiveness of the proposed method on debiasing multi-hop fact verification, we also extend \textbf{PolitiHop} with the help of GPT-4~\cite{openai2023gpt4}.

\textbf{Hard PolitiHop}: We use GPT-4 to generate some misleading evidence sentences for \textbf{PolitiHop}. Specifically, we use GPT-4 to generate modified claims that are similar but parallel to the original claim. GPT-4 then generates evidence for the modified claims that have the opposite label. For example, if the original label is REFUTES, the relation between the generated new evidence and the modified claim should be SUPPORTS. We need to make sure that the Original claim and the new evidence are neither REFUTED nor SUPPORTED. Finally, the new evidence replaces the non-evidence sentences in the original claim.

\textbf{Symmetric PolitiHop}: Previous work~\cite{schuster-etal-2019-towards} has shown that symmetric datasets can evaluate the debiasing ability of models. We combine the new samples generated by GPT-4 and the original samples into the \textbf{Symmetric PolitiHop}. The generated evidence does not change the label of the original claim, so we use the setting of shared evidence (\textbf{ShareEvi}) to increase the difficulty of the dataset. That is, both the new claim and the original claim use the evidence set merged by the new evidence and the original evidence. Here we only consider `SUPPORTS' and `REFUTES' samples.

\textbf{(Adversarial) FEVER-MH}: In FEVER, 83.2\% of the claims require one piece of evidence. Therefore, to evaluate the multi-hop reasoning ability of the model, we extract the samples that have more than 2 pieces of evidence to construct a variant \textbf{FEVER-MH}. We use the same method to construct dataset \textbf{Adversarial FEVER-MH}.

\begin{table*}[]
\centering
\renewcommand\arraystretch{1.1}
\begin{tabular}{cccccccc}
\hline
\textbf{Models} &
  \textbf{PolitiHop} &
  \textbf{\begin{tabular}[c]{@{}c@{}}Adversarial\\ PolitiHop\end{tabular}} &
  \textbf{\begin{tabular}[c]{@{}c@{}}Hard\\ PolitiHop\end{tabular}} &
  \textbf{FEVER} &
  \textbf{FEVER-MH} &
  \textbf{\begin{tabular}[c]{@{}c@{}}Adversarial\\ FEVER\end{tabular}} &
  \textbf{\begin{tabular}[c]{@{}c@{}}Adversarial\\ FEVER-MH\end{tabular}} \\ \hline
BERT-Concat    & 76.00          & 74.50          & 71.50          & 82.14          & 86.32          & 59.08          & 62.12         \\ \hline
GEAR           & 75.50          & 75.00          & 73.50          & 86.58          & 87.04          & 57.81          & 63.36         \\
KGAT           & 77.00          & 74.50          & 74.00          & 86.74          & 89.90          & 59.34          & 64.85         \\ 
Transformer-XH & 75.50          & 77.00          & 72.34          & 83.11          & 86.58          & 59.39          & 64.36         \\
\hline
CICR           & 76.00          & 74.50          & 75.00          & 79.25          & 83.37          & 61.88          & 64.11         \\
CLEVER         & 76.00          & 76.00          & 73.00          & 78.68          & 82.50          & 59.85          & 64.36         \\ \hline
CICR-graph     & 78.00          & 77.50          & 76.50          & 87.38          & 91.53          & 59.21          & 65.84         \\
CLEVER-graph   & 78.00          & 76.50          & 75.50          & 86.24          & 89.99          & 59.47          & 64.60         \\ \hline
Causal Walk    & \textbf{80.00} & \textbf{79.00} & \textbf{79.00} & \textbf{90.19} & \textbf{92.88} & \textbf{62.13} & \textbf{67.08}\\ \hline
\end{tabular}
\caption{Experimental results for PolitiHop dataset, FEVER dataset, and their variants. The best results are in bold.}
\label{tab:main_result}
\end{table*}

\begin{table*}[]
\centering
\renewcommand\arraystretch{1.1}
\begin{tabular}{cccccccc}
\hline
\textbf{Models} &
  \textbf{PolitiHop} &
  \textbf{\begin{tabular}[c]{@{}c@{}}Adversarial\\ PolitiHop\end{tabular}} &
  \textbf{\begin{tabular}[c]{@{}c@{}}Hard\\ PolitiHop\end{tabular}} &
  \textbf{FEVER} &
  \textbf{FEVER-MH} &
  \textbf{\begin{tabular}[c]{@{}c@{}}Adversarial\\ FEVER\end{tabular}} &
  \textbf{\begin{tabular}[c]{@{}c@{}}Adversarial\\ FEVER-MH\end{tabular}} \\ \hline
Causal Walk              & \textbf{80.00} & \textbf{79.00} & \textbf{79.00} & \textbf{90.19} & \textbf{92.88} & \textbf{62.13} & \textbf{67.08} \\ \hline
w/o intervention  & 77.00          & 78.00          & 77.00          & 87.86          & 90.69          & 60.86          & 65.35          \\ \hline
w/ evidence label & 78.00          & 77.00          & 77.00          & 89.48          & 91.87          & 59.72          & 64.36          \\ \hline
\end{tabular}
\caption{Experimental results for ablation study. The best results are in bold.}
\label{tab:ablation}
\end{table*}

\subsection{Baselines}
We compare our proposed method with several baselines including multi-hop fact verification methods and causal-inference-based debiasing methods. 
For multi-hop fact verification methods, \textbf{GEAR}~\cite{zhou-etal-2019-gear}, \textbf{KGAT}~\cite{liu-etal-2020-fine} and \textbf{Transformer-XH}~\cite{zhao2020transformer-xh} are adopted. 
For causal-inference-based debiasing methods for fact verification, \textbf{CICR}~\cite{tian2022debiasing} and \textbf{CLEVER}~\cite{xu-etal-2023-counterfactual} are adopted. 
We also include a baseline for all methods, namely \textbf{BERT-Concat}, where the claim and all evidence are concatenated into a sequence to be fed into the BERT model.
To investigate the effectiveness of previous causal inference-based methods for multi-hop fact verification, we also replace the encoders of \textbf{CICR} and \textbf{CLEVER} with the same graph neural networks of Causal walk: \textbf{CICR-graph} and \textbf{CLEVER-graph}.

\subsection{Implementation Details}
We utilize BERT$_{base}$~\cite{devlin-etal-2019-bert} in all of the BERT fine-tuning baselines and our Causal Walk framework. The learning rate is 1e-5. All models are trained for 10 epochs with a batch size of 4. We update the parameters using Adam optimizer. BERT-Concat, CICR, and CLEVER have a maximum input length of 512, and the other models have a maximum input length of 128. The maximum number $n$ of evidence per sample is 20. The beam width $w$ is 3 and the path sampling length $m$ is 5. The number of samples $k$ for each category in the confounder dictionary is 5. The intervention weight parameter $\alpha$ is 0.1. Following~\cite{xu-etal-2023-counterfactual}, we train models on \textbf{FEVER} data and its variants without using `NOT ENOUGH INFO' samples.

\begin{table}[th]
\centering
\renewcommand\arraystretch{1.0}
\begin{tabular}{cccc}
\hline
\textbf{Models} &
  \textbf{\begin{tabular}[c]{@{}c@{}}PolitiHop\\ -adv\end{tabular}} &
  \textbf{Symmetric} &
  \textbf{\begin{tabular}[c]{@{}c@{}}Symmetric\\ -ShareEvi\end{tabular}} \\ \hline
GEAR &
  \textbf{89.47} &
  51.17 &
  50.88 \\ \hline
CICR-graph   & 87.72          & 51.75          & 50.58          \\
CLEVER-graph & 86.55          & 52.05          & 53.22          \\ \hline
Causal Walk &
  88.30 &
  \textbf{57.02} &
  \textbf{54.09} \\ \hline
\end{tabular}
\caption{Experimental results on Symmetric PolitiHop dataset. The best results are in bold.} 
\label{tab:symmetric}
\end{table}

\subsection{Results}
Table~\ref{tab:main_result} shows the comparison results between our proposed model and other baseline models. It can be observed that our model achieves the best performance on the various variant datasets of PolitiHop and FEVER. On the PolitiHop, each sample contains an average of 4 pieces of evidence and 24 non-evidence sentences, which shows that Causal Walk has better multi-hop reasoning ability. In addition, the performance of Causal Walk is more consistent on the PolitiHop, Adversarial PolitiHop, and Hard PolitiHop. This proves that our model is not confused by misleading evidence and is more robust to data bias. On the FEVER and its variants, the performance of GEAR, KGAT, and Causal Walk on the multi-hop dataset (FEVER-MH) is significantly better than other models, which proves that the graph-based model has strong multi-hop reasoning ability. The performance of CICR, CLEVER, and Causal Walk on Adversarial FEVER data set is significantly better than other models, which proves that the causal-based model has strong robustness. On the multi-hop Adversarial FEVER-MH dataset, our model still has the best performance, proving that our method has all the advantages of both graph-based and causal-based models.

Table~\ref{tab:symmetric} shows the performance of the model on the symmetric dataset. It can be observed that our model achieves the best performance on both Symmetric and Symmetric-ShareEvi.

\subsection{Ablation Study}
To investigate the effectiveness of our proposed causal intervention approach, we evaluate the performance of the model after removing the causal intervention part (w/o intervention), that is, directly using the result $\bm{l}_{pred}$ of Equation (\ref{final_classifier}) as the final classification result. As shown in Table~\ref{tab:ablation}, the performance decreases consistently on both the original dataset and the adversarial dataset, which illustrates the effectiveness of the causal intervention. 

The evidence set contains ground-truth evidence and non-evidence sentences. Both PolitiHop and FEVER provide the evidence labels. During training, we use these evidence labels to supervise the transition probability matrix in Equation (\ref{transition}) (w/ evidence label). Counterintuitively, the performance of the model decreases after using the evidence labels. The performance decreases less on PolitiHop and FEVER, and more on the Adversarial dataset. We believe that this is because the proportion of evidence sentences and non-evidence sentences in the training data is very small, so the transition probability matrix will converge to a very sparse state, which will lead to serious error accumulation in path sampling on the out-of-distribution data set. From another point of view, in the front-door adjustment, we need to sample the paths multiple times and calculate the weighted sum. For the diversity of path sampling, we want the transition probability matrix to be dense rather than sparse.

\section{Conclusion}
In this paper, we propose Causal Walk to debias multi-hop fact verification based on front-door adjustment. 
The reasoning path of the claim-evidence graph is introduced as the mediator between input and output to perform the front-door adjustment.
Specifically, the causal effect between input and output is decomposed into two parts, the causal effect between input and mediator and the causal effect between mediator and output.
The former part is estimated by combining the idea of random walk and graph neural network while the latter part is estimated by introducing NWGM approximation into recurrent neural network.
What's more, we also extend existing datasets with the help of large language models to further test the proposed method.
The experimental results for both existing and new datasets show the effectiveness of the proposed method.

\section{Acknowledgments}
The authors would like to thank the anonymous reviewers for their insightful comments. This work is funded by the National Natural Science Foundation of China (62176053). This work is supported by the Big Data Computing Center of Southeast University.

\bibliography{aaai24}

\begin{thebibliography}{36}
\providecommand{\natexlab}[1]{#1}

\bibitem[{Chen et~al.(2023)Chen, Hu, Li, Shao, and Nie}]{chen-etal-2023-causal}
Chen, Z.; Hu, L.; Li, W.; Shao, Y.; and Nie, L. 2023.
\newblock Causal Intervention and Counterfactual Reasoning for Multi-modal Fake
  News Detection.
\newblock In \emph{Proceedings of the 61st Annual Meeting of the Association
  for Computational Linguistics (Volume 1: Long Papers)}, 627--638. Toronto,
  Canada: Association for Computational Linguistics.

\bibitem[{Chen et~al.(2022)Chen, Hui, Zhuang, Liao, Li, Jia, and
  Li}]{10.1145/3485447.3512135}
Chen, Z.; Hui, S.~C.; Zhuang, F.; Liao, L.; Li, F.; Jia, M.; and Li, J. 2022.
\newblock EvidenceNet: Evidence Fusion Network for Fact Verification.
\newblock In \emph{Proceedings of the ACM Web Conference 2022}, WWW '22,
  2636–2645. New York, NY, USA: Association for Computing Machinery.
\newblock ISBN 9781450390965.

\bibitem[{Christopoulou, Miwa, and
  Ananiadou(2018)}]{christopoulou-etal-2018-walk}
Christopoulou, F.; Miwa, M.; and Ananiadou, S. 2018.
\newblock A Walk-based Model on Entity Graphs for Relation Extraction.
\newblock In \emph{Proceedings of the 56th Annual Meeting of the Association
  for Computational Linguistics (Volume 2: Short Papers)}, 81--88. Melbourne,
  Australia: Association for Computational Linguistics.

\bibitem[{Devlin et~al.(2019)Devlin, Chang, Lee, and
  Toutanova}]{devlin-etal-2019-bert}
Devlin, J.; Chang, M.-W.; Lee, K.; and Toutanova, K. 2019.
\newblock {BERT}: Pre-training of Deep Bidirectional Transformers for Language
  Understanding.
\newblock In \emph{Proceedings of the 2019 Conference of the North {A}merican
  Chapter of the Association for Computational Linguistics: Human Language
  Technologies, Volume 1 (Long and Short Papers)}, 4171--4186. Minneapolis,
  Minnesota: Association for Computational Linguistics.

\bibitem[{Fajcik, Motlicek, and Smrz(2023)}]{fajcik-etal-2023-claim}
Fajcik, M.; Motlicek, P.; and Smrz, P. 2023.
\newblock Claim-Dissector: An Interpretable Fact-Checking System with Joint
  Re-ranking and Veracity Prediction.
\newblock In Rogers, A.; Boyd-Graber, J.; and Okazaki, N., eds., \emph{Findings
  of the Association for Computational Linguistics: ACL 2023}, 10184--10205.
  Toronto, Canada: Association for Computational Linguistics.

\bibitem[{Guo et~al.(2023)Guo, Gong, Rao, and
  Lai}]{guo-etal-2023-counterfactual}
Guo, W.; Gong, Q.; Rao, Y.; and Lai, H. 2023.
\newblock Counterfactual Multihop {QA}: A Cause-Effect Approach for Reducing
  Disconnected Reasoning.
\newblock In \emph{Proceedings of the 61st Annual Meeting of the Association
  for Computational Linguistics (Volume 1: Long Papers)}, 4214--4226. Toronto,
  Canada: Association for Computational Linguistics.

\bibitem[{Hanselowski et~al.(2018)Hanselowski, Zhang, Li, Sorokin, Schiller,
  Schulz, and Gurevych}]{hanselowski-etal-2018-ukp}
Hanselowski, A.; Zhang, H.; Li, Z.; Sorokin, D.; Schiller, B.; Schulz, C.; and
  Gurevych, I. 2018.
\newblock {UKP}-Athene: Multi-Sentence Textual Entailment for Claim
  Verification.
\newblock In \emph{Proceedings of the First Workshop on Fact Extraction and
  {VER}ification ({FEVER})}, 103--108. Brussels, Belgium: Association for
  Computational Linguistics.

\bibitem[{Karimi~Mahabadi, Belinkov, and
  Henderson(2020)}]{karimi-mahabadi-etal-2020-end}
Karimi~Mahabadi, R.; Belinkov, Y.; and Henderson, J. 2020.
\newblock End-to-End Bias Mitigation by Modelling Biases in Corpora.
\newblock In \emph{Proceedings of the 58th Annual Meeting of the Association
  for Computational Linguistics}, 8706--8716. Online: Association for
  Computational Linguistics.

\bibitem[{Lee et~al.(2021)Lee, Won, Kim, Lee, Park, and
  Jung}]{10.1145/3459637.3482078}
Lee, M.; Won, S.; Kim, J.; Lee, H.; Park, C.; and Jung, K. 2021.
\newblock CrossAug: A Contrastive Data Augmentation Method for Debiasing Fact
  Verification Models.
\newblock In \emph{Proceedings of the 30th ACM International Conference on
  Information \& Knowledge Management}, CIKM '21, 3181–3185. New York, NY,
  USA: Association for Computing Machinery.
\newblock ISBN 9781450384469.

\bibitem[{Li, Zhu, and Zhang(2016)}]{li-etal-2016-discriminative}
Li, J.; Zhu, J.; and Zhang, B. 2016.
\newblock Discriminative Deep Random Walk for Network Classification.
\newblock In \emph{Proceedings of the 54th Annual Meeting of the Association
  for Computational Linguistics (Volume 1: Long Papers)}, 1004--1013. Berlin,
  Germany: Association for Computational Linguistics.

\bibitem[{Liu et~al.(2022)Liu, Wang, Yang, Zhou, Yao, Shao, and
  Zhao}]{Liu_2022_CVPR1}
Liu, B.; Wang, D.; Yang, X.; Zhou, Y.; Yao, R.; Shao, Z.; and Zhao, J. 2022.
\newblock Show, Deconfound and Tell: Image Captioning With Causal Inference.
\newblock In \emph{Proceedings of the IEEE/CVF Conference on Computer Vision
  and Pattern Recognition (CVPR)}, 18041--18050.

\bibitem[{Liu et~al.(2020)Liu, Xiong, Sun, and Liu}]{liu-etal-2020-fine}
Liu, Z.; Xiong, C.; Sun, M.; and Liu, Z. 2020.
\newblock Fine-grained Fact Verification with Kernel Graph Attention Network.
\newblock In \emph{Proceedings of the 58th Annual Meeting of the Association
  for Computational Linguistics}, 7342--7351. Online: Association for
  Computational Linguistics.

\bibitem[{Nie, Chen, and Bansal(2019)}]{nie2019combining}
Nie, Y.; Chen, H.; and Bansal, M. 2019.
\newblock Combining fact extraction and verification with neural semantic
  matching networks.
\newblock In \emph{Proceedings of the AAAI conference on artificial
  intelligence}, volume~33, 6859--6866.

\bibitem[{Niu et~al.(2021)Niu, Tang, Zhang, Lu, Hua, and
  Wen}]{niu2021counterfactual}
Niu, Y.; Tang, K.; Zhang, H.; Lu, Z.; Hua, X.-S.; and Wen, J.-R. 2021.
\newblock Counterfactual vqa: A cause-effect look at language bias.
\newblock In \emph{Proceedings of the IEEE/CVF Conference on Computer Vision
  and Pattern Recognition}, 12700--12710.

\bibitem[{OpenAI(2023)}]{openai2023gpt4}
OpenAI. 2023.
\newblock GPT-4 Technical Report.
\newblock \url{https://openai.com/research/gpt-4}.
\newblock Accessed: 2023-07-28, arXiv:2303.08774.

\bibitem[{Ostrowski et~al.(2021)Ostrowski, Arora, Atanasova, and
  Augenstein}]{ijcai2021p536}
Ostrowski, W.; Arora, A.; Atanasova, P.; and Augenstein, I. 2021.
\newblock Multi-Hop Fact Checking of Political Claims.
\newblock In Zhou, Z.-H., ed., \emph{Proceedings of the Thirtieth International
  Joint Conference on Artificial Intelligence, {IJCAI-21}}, 3892--3898.
  International Joint Conferences on Artificial Intelligence Organization.
\newblock Main Track.

\bibitem[{Pan et~al.(2023)Pan, Wu, Lu, Luu, Wang, Kan, and
  Nakov}]{pan-etal-2023-fact}
Pan, L.; Wu, X.; Lu, X.; Luu, A.~T.; Wang, W.~Y.; Kan, M.-Y.; and Nakov, P.
  2023.
\newblock Fact-Checking Complex Claims with Program-Guided Reasoning.
\newblock In Rogers, A.; Boyd-Graber, J.; and Okazaki, N., eds.,
  \emph{Proceedings of the 61st Annual Meeting of the Association for
  Computational Linguistics (Volume 1: Long Papers)}, 6981--7004. Toronto,
  Canada: Association for Computational Linguistics.

\bibitem[{Schuster et~al.(2019)Schuster, Shah, Yeo, Roberto Filizzola~Ortiz,
  Santus, and Barzilay}]{schuster-etal-2019-towards}
Schuster, T.; Shah, D.; Yeo, Y. J.~S.; Roberto Filizzola~Ortiz, D.; Santus, E.;
  and Barzilay, R. 2019.
\newblock Towards Debiasing Fact Verification Models.
\newblock In \emph{Proceedings of the 2019 Conference on Empirical Methods in
  Natural Language Processing and the 9th International Joint Conference on
  Natural Language Processing (EMNLP-IJCNLP)}, 3419--3425. Hong Kong, China:
  Association for Computational Linguistics.

\bibitem[{Si, Zhu, and Zhou(2023)}]{si2023exploring}
Si, J.; Zhu, Y.; and Zhou, D. 2023.
\newblock Exploring faithful rationale for multi-hop fact verification via
  salience-aware graph learning.
\newblock In \emph{Proceedings of the AAAI Conference on Artificial
  Intelligence}, volume~37, 13573--13581.

\bibitem[{Thorne et~al.(2018)Thorne, Vlachos, Christodoulopoulos, and
  Mittal}]{thorne-etal-2018-fever}
Thorne, J.; Vlachos, A.; Christodoulopoulos, C.; and Mittal, A. 2018.
\newblock {FEVER}: a Large-scale Dataset for Fact Extraction and
  {VER}ification.
\newblock In \emph{Proceedings of the 2018 Conference of the North {A}merican
  Chapter of the Association for Computational Linguistics: Human Language
  Technologies, Volume 1 (Long Papers)}, 809--819. New Orleans, Louisiana:
  Association for Computational Linguistics.

\bibitem[{Tian et~al.(2022{\natexlab{a}})Tian, Cao, Zhang, and
  Xing}]{Tian_Cao_Zhang_Xing_2022}
Tian, B.; Cao, Y.; Zhang, Y.; and Xing, C. 2022{\natexlab{a}}.
\newblock Debiasing NLU Models via Causal Intervention and Counterfactual
  Reasoning.
\newblock \emph{Proceedings of the AAAI Conference on Artificial Intelligence},
  36(10): 11376–11384.

\bibitem[{Tian et~al.(2022{\natexlab{b}})Tian, Cao, Zhang, and
  Xing}]{tian2022debiasing}
Tian, B.; Cao, Y.; Zhang, Y.; and Xing, C. 2022{\natexlab{b}}.
\newblock Debiasing NLU models via causal intervention and counterfactual
  reasoning.
\newblock In \emph{Proceedings of the AAAI Conference on Artificial
  Intelligence}, volume~36, 11376--11384.

\bibitem[{Trivedi et~al.(2020)Trivedi, Balasubramanian, Khot, and
  Sabharwal}]{trivedi-etal-2020-multihop}
Trivedi, H.; Balasubramanian, N.; Khot, T.; and Sabharwal, A. 2020.
\newblock Is Multihop {QA} in {DiRe} Condition? Measuring and Reducing
  Disconnected Reasoning.
\newblock In \emph{Proceedings of the 2020 Conference on Empirical Methods in
  Natural Language Processing (EMNLP)}, 8846--8863. Online: Association for
  Computational Linguistics.

\bibitem[{Vaswani et~al.(2017)Vaswani, Shazeer, Parmar, Uszkoreit, Jones,
  Gomez, Kaiser, and Polosukhin}]{NIPS2017_3f5ee243}
Vaswani, A.; Shazeer, N.; Parmar, N.; Uszkoreit, J.; Jones, L.; Gomez, A.~N.;
  Kaiser, L.~u.; and Polosukhin, I. 2017.
\newblock Attention is All you Need.
\newblock In Guyon, I.; Luxburg, U.~V.; Bengio, S.; Wallach, H.; Fergus, R.;
  Vishwanathan, S.; and Garnett, R., eds., \emph{Advances in Neural Information
  Processing Systems}, volume~30. Curran Associates, Inc.

\bibitem[{Wang et~al.(2022)Wang, Zhou, Sun, Ye, Gui, Zhang, and
  Huang}]{wang-etal-2022-causal}
Wang, S.; Zhou, J.; Sun, C.; Ye, J.; Gui, T.; Zhang, Q.; and Huang, X. 2022.
\newblock Causal Intervention Improves Implicit Sentiment Analysis.
\newblock In \emph{Proceedings of the 29th International Conference on
  Computational Linguistics}, 6966--6977. Gyeongju, Republic of Korea:
  International Committee on Computational Linguistics.

\bibitem[{Wei and Zou(2019)}]{wei-zou-2019-eda}
Wei, J.; and Zou, K. 2019.
\newblock {EDA}: Easy Data Augmentation Techniques for Boosting Performance on
  Text Classification Tasks.
\newblock In \emph{Proceedings of the 2019 Conference on Empirical Methods in
  Natural Language Processing and the 9th International Joint Conference on
  Natural Language Processing (EMNLP-IJCNLP)}, 6382--6388. Hong Kong, China:
  Association for Computational Linguistics.

\bibitem[{Xu et~al.(2015)Xu, Ba, Kiros, Cho, Courville, Salakhudinov, Zemel,
  and Bengio}]{Xu_Ba_Kiros_Cho_Courville_Salakhudinov_Zemel_Bengio_2015}
Xu, K.; Ba, J.; Kiros, R.; Cho, K.; Courville, A.; Salakhudinov, R.; Zemel, R.;
  and Bengio, Y. 2015.
\newblock Show, Attend and Tell: Neural Image Caption Generation with Visual
  Attention.
\newblock \emph{International Conference on Machine Learning,International
  Conference on Machine Learning}.

\bibitem[{Xu et~al.(2023)Xu, Liu, Wu, and Wang}]{xu-etal-2023-counterfactual}
Xu, W.; Liu, Q.; Wu, S.; and Wang, L. 2023.
\newblock Counterfactual Debiasing for Fact Verification.
\newblock In \emph{Proceedings of the 61st Annual Meeting of the Association
  for Computational Linguistics (Volume 1: Long Papers)}, 6777--6789. Toronto,
  Canada: Association for Computational Linguistics.

\bibitem[{Yang, Zhang, and Cai(2021)}]{yang2021deconfounded}
Yang, X.; Zhang, H.; and Cai, J. 2021.
\newblock Deconfounded image captioning: A causal retrospect.
\newblock \emph{IEEE Transactions on Pattern Analysis and Machine
  Intelligence}.

\bibitem[{Yang et~al.(2021)Yang, Zhang, Qi, and Cai}]{yang2021causal}
Yang, X.; Zhang, H.; Qi, G.; and Cai, J. 2021.
\newblock Causal attention for vision-language tasks.
\newblock In \emph{Proceedings of the IEEE/CVF conference on computer vision
  and pattern recognition}, 9847--9857.

\bibitem[{Zeng and Gao(2023)}]{zeng-gao-2023-prompt}
Zeng, F.; and Gao, W. 2023.
\newblock Prompt to be Consistent is Better than Self-Consistent? Few-Shot and
  Zero-Shot Fact Verification with Pre-trained Language Models.
\newblock In Rogers, A.; Boyd-Graber, J.; and Okazaki, N., eds., \emph{Findings
  of the Association for Computational Linguistics: ACL 2023}, 4555--4569.
  Toronto, Canada: Association for Computational Linguistics.

\bibitem[{Zhao et~al.(2020{\natexlab{a}})Zhao, Xiong, Rosset, Song, Bennett,
  and Tiwary}]{Zhao2020Transformer-XH:}
Zhao, C.; Xiong, C.; Rosset, C.; Song, X.; Bennett, P.; and Tiwary, S.
  2020{\natexlab{a}}.
\newblock Transformer-XH: Multi-Evidence Reasoning with eXtra Hop Attention.
\newblock In \emph{International Conference on Learning Representations}.

\bibitem[{Zhao et~al.(2020{\natexlab{b}})Zhao, Xiong, Rosset, Song, Bennett,
  and Tiwary}]{zhao2020transformer-xh}
Zhao, C.; Xiong, C.; Rosset, C.; Song, X.; Bennett, P.; and Tiwary, S.
  2020{\natexlab{b}}.
\newblock Transformer-XH: Multi-evidence Reasoning with Extra Hop Attention.
\newblock In \emph{The Eighth International Conference on Learning
  Representations (ICLR 2020)}.

\bibitem[{Zhong et~al.(2020)Zhong, Xu, Tang, Xu, Duan, Zhou, Wang, and
  Yin}]{zhong-etal-2020-reasoning}
Zhong, W.; Xu, J.; Tang, D.; Xu, Z.; Duan, N.; Zhou, M.; Wang, J.; and Yin, J.
  2020.
\newblock Reasoning Over Semantic-Level Graph for Fact Checking.
\newblock In \emph{Proceedings of the 58th Annual Meeting of the Association
  for Computational Linguistics}, 6170--6180. Online: Association for
  Computational Linguistics.

\bibitem[{Zhou et~al.(2019)Zhou, Han, Yang, Liu, Wang, Li, and
  Sun}]{zhou-etal-2019-gear}
Zhou, J.; Han, X.; Yang, C.; Liu, Z.; Wang, L.; Li, C.; and Sun, M. 2019.
\newblock {GEAR}: Graph-based Evidence Aggregating and Reasoning for Fact
  Verification.
\newblock In \emph{Proceedings of the 57th Annual Meeting of the Association
  for Computational Linguistics}, 892--901. Florence, Italy: Association for
  Computational Linguistics.

\bibitem[{Zhu et~al.(2023)Zhu, Wu, Zhang, Hou, and Feng}]{zhu-etal-2023-causal}
Zhu, J.; Wu, S.; Zhang, X.; Hou, Y.; and Feng, Z. 2023.
\newblock Causal Intervention for Mitigating Name Bias in Machine Reading
  Comprehension.
\newblock In \emph{Findings of the Association for Computational Linguistics:
  ACL 2023}, 12837--12852. Toronto, Canada: Association for Computational
  Linguistics.

\end{thebibliography}

\end{document}